\documentclass{article}
\usepackage[numbers]{natbib}
\usepackage[margin=1in]{geometry}   
\usepackage{times}                  
\usepackage{microtype}              
\usepackage[utf8]{inputenc}
\usepackage[T1]{fontenc}

\usepackage{amsmath, amssymb, amsfonts}

\usepackage{graphicx}
\usepackage{subcaption}

\usepackage{booktabs}
\usepackage{multirow}

\usepackage{xcolor}
\usepackage{nicefrac}
\usepackage{microtype}

\usepackage{algorithm}
\usepackage{algpseudocode}

\usepackage{float}

\usepackage{url}
\usepackage{hyperref}

\title{
Feasibility of Simulated Adversarial Patch Attacks\\
on Cargo Occupancy Estimation Systems
}

\author{
  Sesugh Nder \\
  Department of Informatics \\
  University of Hamburg \\
  Vogt-K{\"o}lln-Stra{\ss}e 30, 22527 \\
  \texttt{sesugh.nder@studium.uni-hamburg.de}
  \and
  Mohamed Rissal Hedna \\
  Department of Informatics \\
  University of Hamburg \\
  Vogt-K{\"o}lln-Stra{\ss}e 30, 22527 \\
  \texttt{mohamed.rissal.hedna@studium.uni-hamburg.de}
}

\begin{document}
\maketitle

\begin{abstract}
Computer vision systems are increasingly adopted in modern logistics operations, including the estimation of trailer occupancy for planning, routing, and billing. Although effective, such systems may be vulnerable to physical adversarial attacks, particularly adversarial patches that can be printed and placed on interior surfaces. In this work, we study the feasibility of such attacks on a convolutional cargo-occupancy classifier using fully simulated 3D environments. Using Mitsuba~3 for differentiable rendering, we optimize patch textures across variations in geometry, lighting, and viewpoint, and compare their effectiveness to a 2D compositing baseline. Our experiments demonstrate that 3D-optimized patches achieve high attack success rates, especially in a denial-of-service scenario (empty $\to$ full), where success reaches $84.94\%$. Concealment attacks (full $\to$ empty) prove more challenging but still reach $30.32\%$. We analyze the factors influencing attack success, discuss implications for the security of automated logistics pipelines, and highlight directions for strengthening physical robustness. To our knowledge, this is the first study to investigate adversarial patch attacks for cargo-occupancy estimation in physically realistic, fully simulated 3D scenes.
\end{abstract}

\section{Introduction}

Computer vision systems have become integral to modern logistics, supporting tasks such as trailer occupancy estimation, volumetric analysis, automated inspection, and digital twins of warehouse operations. As these systems increasingly influence routing, billing, and operational decision-making, their robustness becomes a critical safety and security concern.

Recent advances in adversarial machine learning have shown that physical artifacts---such as printed stickers or posters---can cause classifiers to systematically misbehave. Adversarial patches, in particular, have been demonstrated to fool models in traffic sign classification \cite{brown2017adversarial, eykholt2018robust}, surveillance \cite{thys2019foolingautomatedsurveillancecameras}, and other industrial domains. Yet, little is known about their effectiveness in logistically realistic environments such as cargo trailers, where lighting, geometry, and surface textures differ substantially from typical benchmarks.

\textbf{Research question.}
We therefore ask:
\begin{quote}
    \emph{Can adversarial patches designed entirely in simulation reliably mislead a cargo occupancy classifier?}
\end{quote}

Such an attack could enable:
\begin{itemize}
    \item \textbf{Denial-of-service (DoS):} making an empty trailer appear full, potentially interfering with scheduling and billing.
    \item \textbf{Concealment:} making a full trailer appear empty, facilitating theft or bypassing inspection systems.
\end{itemize}

Cargo trailers present unique obstacles for adversarial attacks: lighting varies widely, interior surfaces can be reflective or low-texture, cargo geometry creates complex occlusions, and cameras are mounted far from the patch, leading to steep viewing angles. To address these challenges, we develop a differentiable 3D patch optimization framework based on Mitsuba~3.

\begin{figure}[H]
    \centering
    \includegraphics[width=0.9\linewidth]{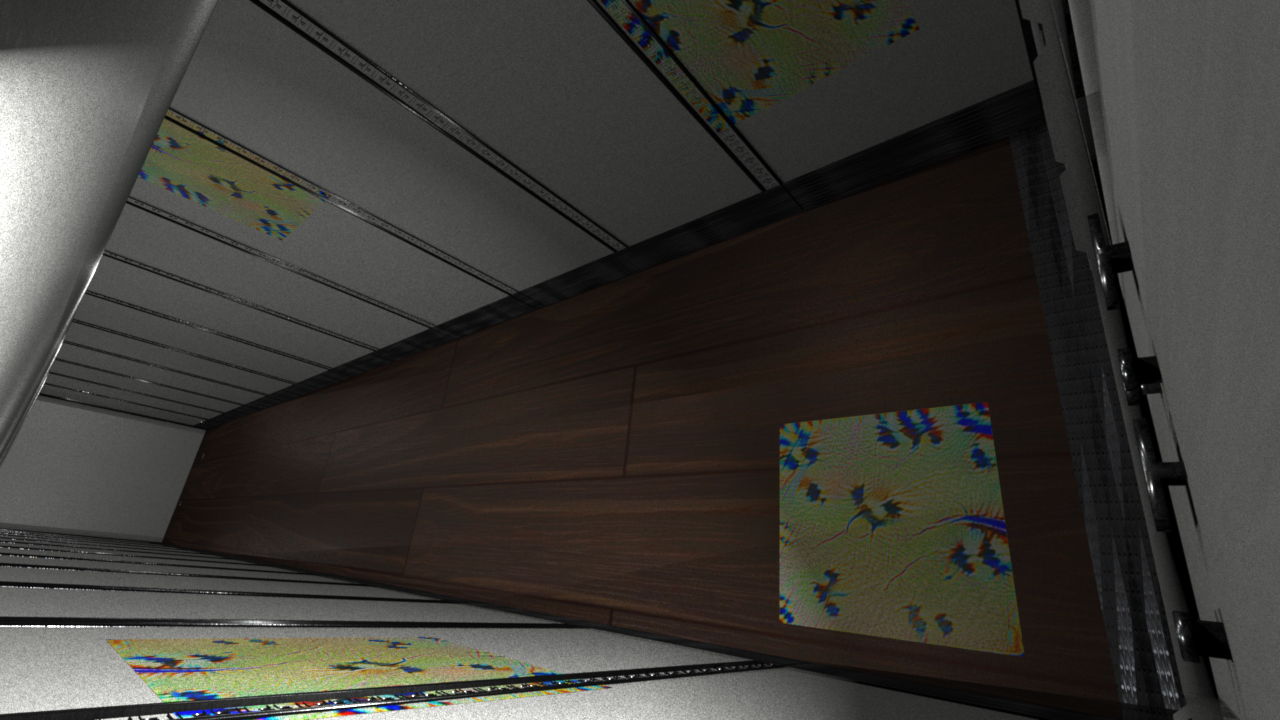}
    \caption{Illustrative example of a successful adversarial patch attack in our simulated cargo trailer environment. A learned patch placed inside the trailer changes the prediction of a high-accuracy occupancy classifier.}
    \label{fig:intro_example}
\end{figure}

\paragraph{Contributions.}
Our main contributions are:
\begin{enumerate}
    \item A differentiable 3D simulation pipeline for adversarial patch optimization in cargo trailers, including surface-aware placement sampling and Mitsuba~3-based rendering.
    \item A comparison of 2D image-space and 3D scene-space patch optimization on 5{,}910 fully rendered scenes, highlighting the benefits of physical realism.
    \item A systematic evaluation of concealment and denial-of-service attacks, revealing strong vulnerabilities in a modern occupancy classifier.
    \item An analysis of lighting, geometry, and patch visibility effects on attack robustness in a logistics-relevant setting.
\end{enumerate}

\section{Related Work}

\paragraph{Adversarial patches.}
Physical adversarial patches were introduced by Brown et al.\ \cite{brown2017adversarial} as universal, localized perturbations capable of inducing targeted misclassifications. Subsequent work has improved physical robustness under viewpoint and lighting changes \cite{eykholt2018robust, thys2019foolingautomatedsurveillancecameras}. Naturalistic patch generation using GANs \cite{hu2021naturalistic} and diffusion models \cite{wang2024diffpatch} seeks to increase stealth by making patches resemble benign textures.

\paragraph{Physical robustness and EOT.}
The Expectation over Transformation (EOT) framework \cite{pmlr-v80-athalye18b} optimizes perturbations over a distribution of transformations (e.g., scaling, rotation, lighting), improving transfer to the real world. The APRICOT dataset \cite{apricot} and subsequent analyses such as \cite{shack2024breakingillusionrealworldchallenges} highlight the challenges of maintaining patch performance across diverse environments.

\paragraph{Differentiable rendering attacks.}
Differentiable renderers such as PyTorch3D \cite{ravi2020pytorch3d} and Mitsuba~3 enable end-to-end adversarial optimization in 3D environments. Prior work has explored digital-to-physical transfer \cite{liu2023digital2physical} and attacks in remote sensing \cite{Digital-to-Physical}. Our work extends this line of research to the logistics domain, focusing on trailer occupancy estimation.

\paragraph{Cargo occupancy estimation.}
Vision-based trailer occupancy estimation has seen initial industrial deployment \cite{kwak2022volume}, enabling automated volume estimation and capacity planning. However, its robustness to adversarial manipulation remains largely unexplored. Our study addresses this gap using a simulation-driven analysis.

\section{Method}

We optimize a learnable patch texture \( p \in \mathbb{R}^{H \times W \times 3} \) that is inserted into 3D scenes via a differentiable renderer and evaluated by a fixed occupancy classifier.

\subsection{Threat Model}

We assume an adversary who:
\begin{itemize}
    \item can print arbitrary patch textures (e.g., stickers or posters),
    \item can place patches on interior surfaces of a trailer (walls, floor, cargo),
    \item has white-box access to the classifier (model architecture and weights),
    \item cannot change the camera pose, scene geometry, or model training data.
\end{itemize}

The adversary seeks to design a patch that induces either targeted or untargeted misclassification under realistic physical variability.

\subsection{Surface-Based Patch Placement}

Instead of directly optimizing patch position, we use a surface-based sampling strategy:

\begin{enumerate}
    \item Cast dense rays from the camera into the interior of the trailer.
    \item Collect visible ray--surface intersection points.
    \item Filter intersections by geometric criteria (e.g., minimum surface area, non-grazing angles).
    \item Sample a valid intersection as a patch anchor location.
    \item Project the patch as a planar surface at this 3D location.
\end{enumerate}

This procedure ensures that patches are placed only at physically plausible, visible locations within the scene.

\subsection{Differentiable Rendering Optimization}

For each optimization step, we:
\begin{enumerate}
    \item sample a trailer scene \(s\) with a given occupancy label,
    \item sample a valid patch placement \(r\) in that scene,
    \item insert the patch \(p\) at \(r\),
    \item render a color image \(x' = \text{Render}(s, p, r)\) using Mitsuba~3,
    \item pass \(x'\) through the classifier to obtain logits \(z = f(x')\),
    \item compute a loss and backpropagate gradients to update the patch texture.
\end{enumerate}

For a targeted attack towards target label \(y_t\), we minimize
\[
\mathcal{L}_\text{target}(x') = -\log z_{y_t},
\]
while for an untargeted attack we maximize the loss associated with the true label \(y\),
\[
\mathcal{L}_\text{untargeted}(x') = \log z_{y}.
\]

\begin{algorithm}[H]
\caption{3D Adversarial Patch Optimization via Differentiable Rendering}
\label{alg:patch_optimization}
\begin{algorithmic}[1]
\Require Scenes $\mathcal{S}$, classifier $f$, initial patch $p$, iterations $T$, learning rate $\eta$, target label $y_t$ (optional)
\For{$t = 1$ to $T$}
    \State Sample scene $s \sim \mathcal{S}$ with label $y$
    \State Compute ray--surface intersections $\mathcal{R}(s)$
    \State Sample placement $r \sim \mathcal{R}(s)$
    \State Insert patch $p$ at $r$ in scene $s$
    \State $x' \gets \text{Render}(s, p, r)$
    \State $z \gets f(x')$
    \If{targeted}
        \State $\mathcal{L} \gets -\log z_{y_t}$
    \Else
        \State $\mathcal{L} \gets \log z_{y}$
    \EndIf
    \State Update patch: $p \gets p - \eta \nabla_{p} \mathcal{L}$
\EndFor
\State \textbf{return} optimized patch $p$
\end{algorithmic}
\end{algorithm}

\begin{figure}[H]
    \centering
    \includegraphics[width=\linewidth]{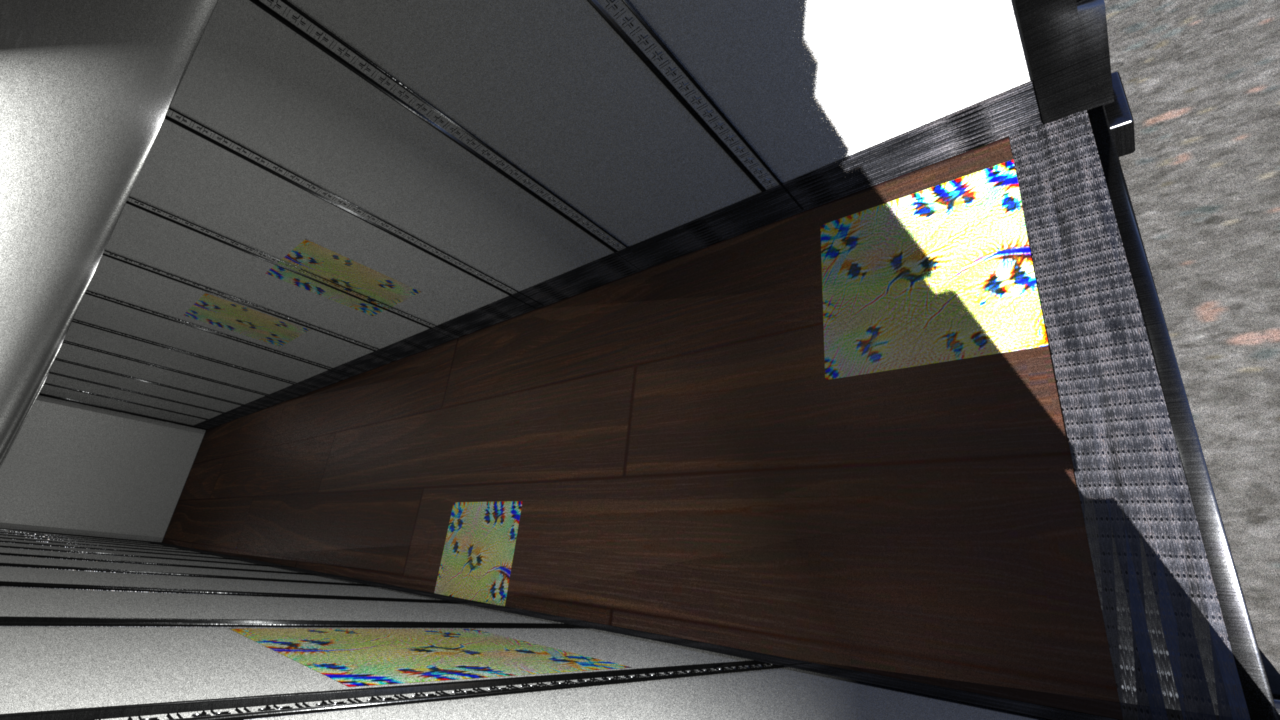}
    \caption{Example of a learned patch rendered under different lighting conditions. Differentiable rendering through Mitsuba~3 allows us to optimize patch textures across illumination variations, improving physical plausibility.}
    \label{fig:lighting_robustness}
\end{figure}

\subsection{2D Image-Space Baseline}

For comparison, we implement a 2D baseline where a clean background image is rendered once, and the patch is composited in image space using a homography induced by the 3D placement. This approach is computationally cheaper but does not model shadows, indirect illumination, or occlusions, and thus lacks full physical realism.

\section{Experiments}

\subsection{Dataset}

We generate 5{,}910 synthetic Mitsuba~3 scenes of a 53-foot cargo trailer with three occupancy classes: empty, medium, full. Scenes vary in lighting, camera pose, and cargo configuration.

\begin{table}[H]
\centering
\caption{Dataset distribution across training and test splits.}
\begin{tabular}{lcc}
\toprule
Class & Train & Test \\
\midrule
Empty  & 1383 & 346 \\
Medium & 2215 & 554 \\
Full   & 1129 & 283 \\
\bottomrule
\end{tabular}
\label{tab:dataset}
\end{table}

\subsection{Victim Model}

We use ConvNeXtV2-Atto \cite{Woo2023ConvNeXtV2} as the victim model, a compact yet high-performing convolutional architecture. The model is initialized from ImageNet pretraining and fine-tuned on our synthetic dataset. On the held-out test set, it achieves near-perfect performance across all three occupancy classes.

\begin{figure}[H]
    \centering
    \includegraphics[width=0.7\linewidth]{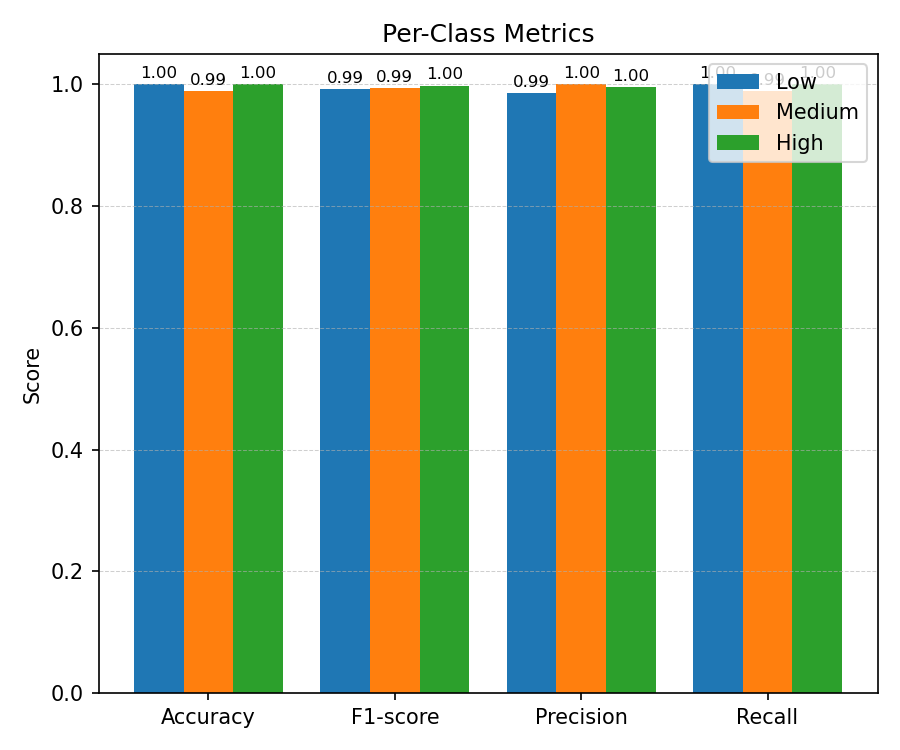}
    \caption{Per-class performance of the victim model on clean test images. Accuracy, precision, recall, and F1-score are all above 0.99 for each occupancy class.}
    \label{fig:victim_metrics}
\end{figure}

\subsection{Evaluation Metrics}

We report:
\begin{itemize}
    \item \textbf{Attack Success Rate (ASR):} percentage of inputs for which the attack achieves the desired misclassification.
    \item \textbf{Overall accuracy:} accuracy of the classifier under attack (for reference).
\end{itemize}

We focus on two targeted attack scenarios:
\begin{itemize}
    \item \textbf{Denial-of-Service (DoS):} empty $\to$ full.
    \item \textbf{Concealment:} full $\to$ empty.
\end{itemize}

\subsection{Attack Success Rates}

Table~\ref{tab:asr_results} summarizes ASR for 2D and 3D optimization.

\begin{table}[H]
\centering
\caption{Attack Success Rates (ASR) for different optimization environments and scenarios.}
\label{tab:asr_results}
\begin{tabular}{lll}
\toprule
\textbf{Environment} & \textbf{Scenario} & \textbf{ASR (\%)} \\
\midrule
2D  & Concealment       & 10.11 \\
3D  & Concealment       & \textbf{30.32} \\
2D  & Denial-of-Service & 26.51 \\
3D  & Denial-of-Service & \textbf{84.94} \\
\bottomrule
\end{tabular}
\end{table}

\subsection{Qualitative Results}

\paragraph{Denial-of-Service (empty $\to$ full).}

\begin{figure}[H]
    \centering
    \includegraphics[width=\linewidth]{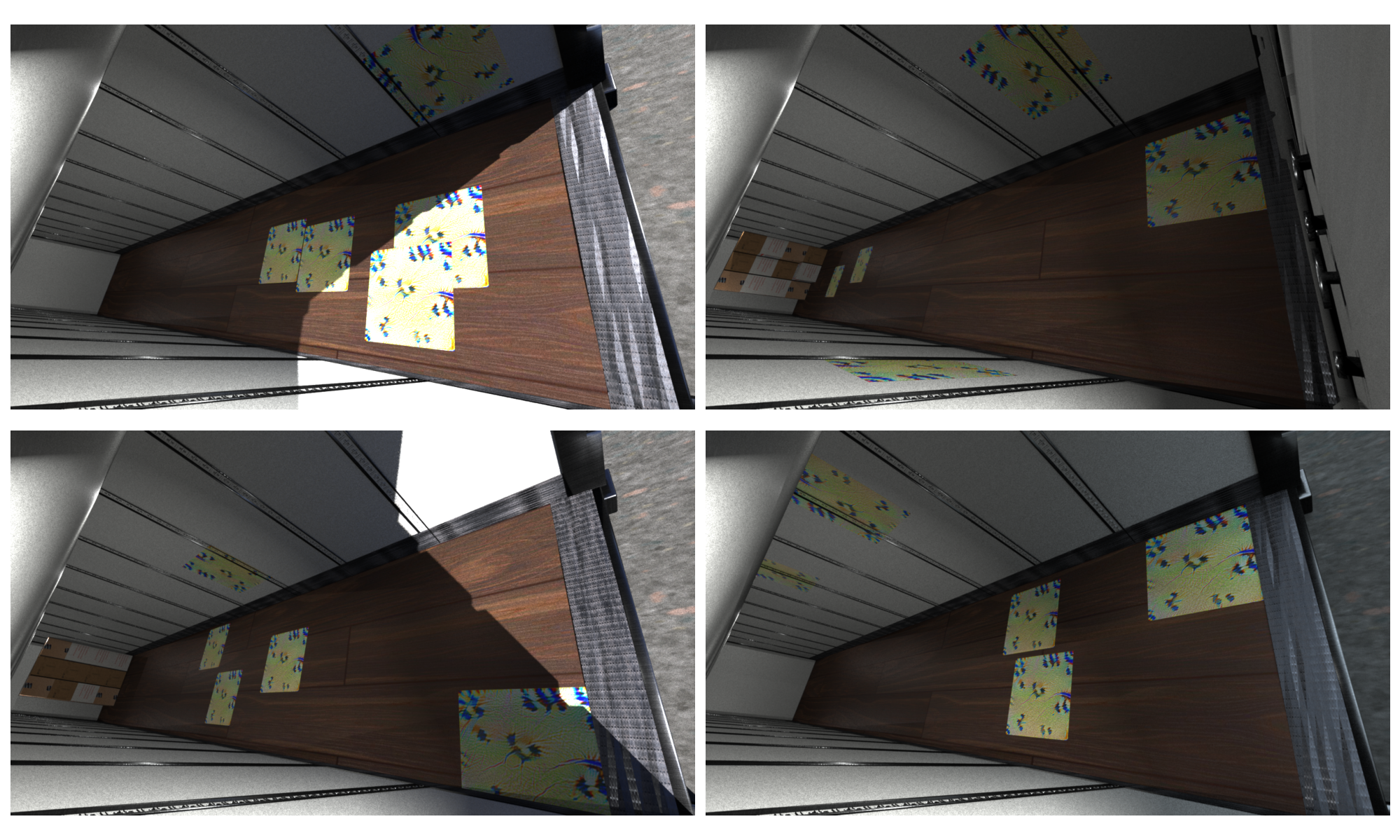}
    \caption{Examples of successful Denial-of-Service attacks (empty $\to$ full) using 3D-optimized patches. The patches are integrated into the scene with realistic lighting and occlusion, leading the classifier to predict a full trailer.}
    \label{fig:dos_3d_examples}
\end{figure}

The Denial-of-Service attack achieves an ASR of 84.94\% when optimized in 3D. Patches placed on the floor or walls introduce strong cues that mimic fully loaded trailers, making it relatively easy for the classifier to overestimate occupancy.

\paragraph{Concealment (full $\to$ empty).}

\begin{figure}[H]
    \centering
    \includegraphics[width=\linewidth]{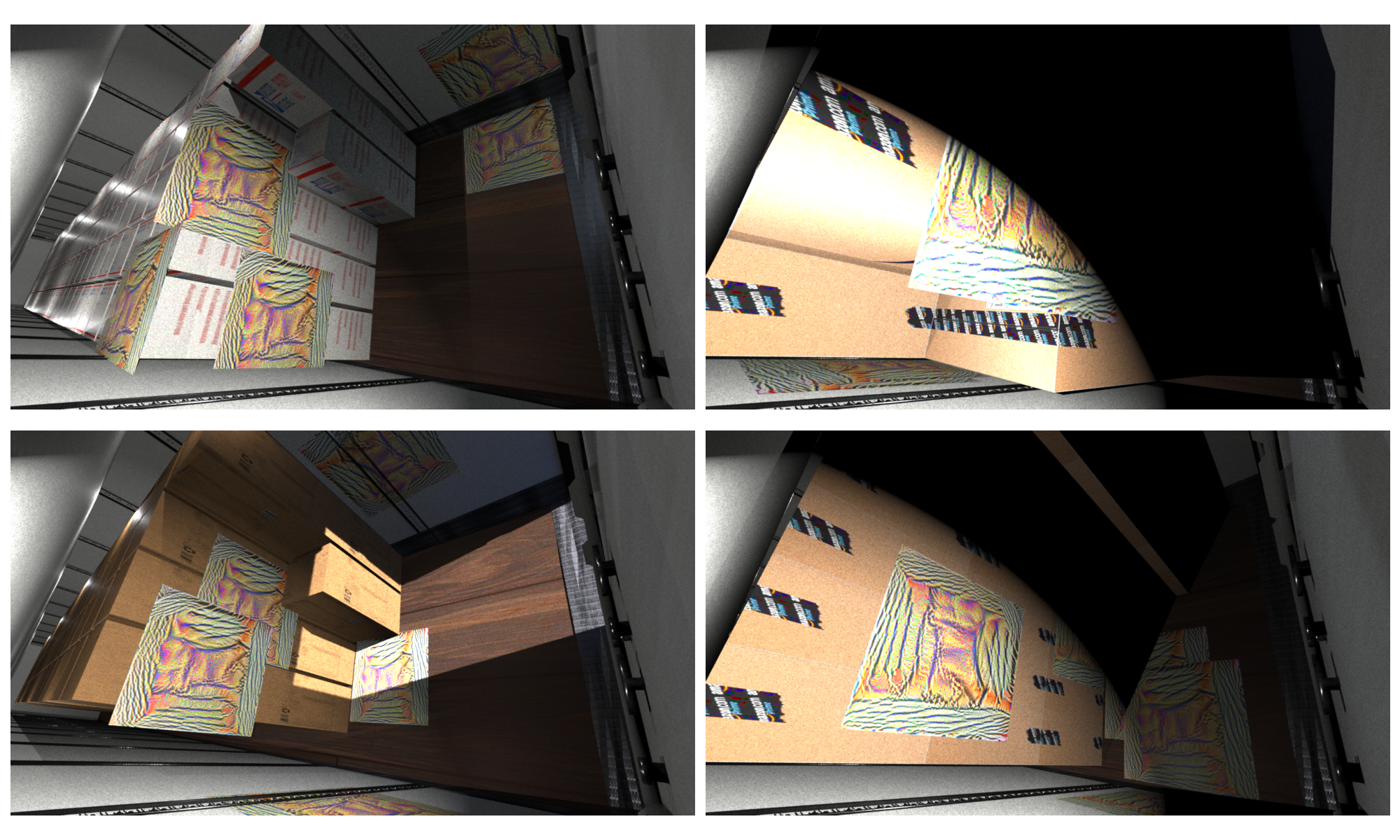}
    \caption{Examples of successful Concealment attacks (full $\to$ empty) in the 3D-rendered environment. Patches placed directly on cargo surfaces interfere with the visual features used by the classifier, causing it to misclassify full trailers as empty.}
    \label{fig:concealment_3d_examples}
\end{figure}

Concealment is more challenging, with an ASR of 30.32\%. The classifier's internal representation of ``full'' scenes appears more robust, as it is supported by strong geometric and textural evidence from the cargo itself.

\section{Discussion}

Our experiments lead to several observations:

\begin{itemize}
    \item \textbf{Asymmetry of decision boundaries.} It is substantially easier to push empty scenes into the ``full'' class than to make full scenes appear empty. This suggests that the decision boundary around the empty class is more easily perturbed.
    \item \textbf{Importance of physical realism.} 3D-optimized patches significantly outperform their 2D counterparts, highlighting that modeling lighting, occlusions, and true 3D geometry is crucial for strong physical attacks.
    \item \textbf{Visibility vs.\ stealth.} Larger, high-contrast patches and multiple placements tend to yield higher ASR but are less likely to remain unnoticed by human inspectors. Balancing attack success with stealth remains an open problem.
\end{itemize}

These findings suggest that current occupancy models may over-rely on global texture statistics and are not robust to structured, localized perturbations introduced in the physical scene.

\section{Conclusion}

We presented a simulation-driven study of adversarial patch attacks on cargo occupancy estimation systems. Combining a differentiable renderer (Mitsuba~3) with surface-based patch placement, we showed that physically plausible patches can reliably fool a high-performing ConvNeXtV2-Atto classifier, particularly in a denial-of-service scenario.

Our work emphasizes the need for adversarially robust perception in logistics automation. Future work includes physical-world validation with printed patches, multi-view and video-based attacks, and defenses such as adversarial training, uncertainty estimation, and multi-sensor fusion.

\clearpage  

\bibliographystyle{plain}
\bibliography{bib}

\end{document}